\documentclass[letterpaper, 10 pt, conference]{ieeeconf}  % Comment this line out if you need a4paper

\IEEEoverridecommandlockouts                              % This command is only needed if
\usepackage{amsfonts,amssymb,mathrsfs,graphicx,amsmath,balance,color,enumerate,booktabs, subfigure}

\usepackage{amsthm}
\newtheoremstyle{mystyle} % style name
            {3pt}         % vertical space above
            {3pt}         % vertical space below
            {\normalfont} % body font
            {}            % indent dim
            {\bfseries}   % Theorem head font
            {:}           % punctuation after theorem head
            {0.5em}       % space after theorem head
            {}            % theorem spec
\theoremstyle{mystyle}

\usepackage{varwidth}
\usepackage{algorithm}
\usepackage{algpseudocode}
\makeatletter
\newcommand{\algmargin}{\the\ALG@thistlm}
\makeatother
\newlength{\whilewidth}
\algdef{SE}[parWHILE]{parWhile}{EndparWhile}[1]
  {\parbox[t]{\dimexpr\linewidth-\algmargin}{%
     \hangindent\whilewidth\strut\algorithmicwhile\ #1\ \algorithmicdo\strut}}{\algorithmicend\ \algorithmicwhile}%
\algnewcommand{\parState}[1]{\State%
  \parbox[t]{\dimexpr\linewidth-\algmargin}{\strut #1\strut}}

\usepackage{bm}
\usepackage[colorlinks,linkcolor=black,urlcolor=blue]{hyperref}
\title{\LARGE \bf
Reinforcement Learning based Negotiation-aware Motion Planning of {Autonomous} Vehicles
}

\author{
    {Zhitao Wang*},
    {Yuzheng Zhuang*},
    Qiang Gu,
    {Dong Chen},
    {Hongbo Zhang},
    Wulong Liu
\thanks{ {* denotes for equally contribution to this paper.}
Z. Wang, Y. Zhuang, Q. Gu, {D. Chen, H, Zhang} and W. Liu are all with Noah's Ark Lab, Huawei Technologies, Beijing, China. (\{wangzhitao5; zhuangyuzheng; qiang.gu; chendong48; zhanghongbo888; liuwulong\}@huawei.com).}
}

\begin{document}

\maketitle
\thispagestyle{empty}
\pagestyle{empty}
\iffalse
\bibliography{ref/refs}
\fi

%%%%%%%%%%%%%%%%%%%%%%%%%%%%%%%%%%%%%%%%%%%%%%%%%%%%%%%%%%%%%%%%%%%%%%%%%%%%%%%%
\begin{abstract}

% This paper proposes a RL-based negotiation-aware motion planning framework for interaction decision making of autonomous vehicles, which combines reinforcement learning and model-based planning. The negotiation-aware framework adopts RL to adjust the planning behavior adaptively according to dynamic environment by modifing the parameter of model-based planner online. Through interacting with social vehicles and regulating planning behavior online, our method could alleviate the social dilemma problem. Towards this goal, a temporal sequence of occupancy grid maps (OGM) is taken as input for RL stack to embed intention reasoning in our framework. Curriculum learning is employed to enhance the training efficiency. We applied our method to narrow lane driving, a typical interaction scenario in both simulation and real-world driving to evaluate the effectiveness of our framework. The strength of our framework is illustrated by a comparison with polynomial planner.

For autonomous vehicles integrating onto roadways with human traffic participants, it requires understanding and adapting to the participants' intention and driving styles by responding in predictable ways without explicit communication. 
This paper proposes a reinforcement learning based negotiation-aware motion planning framework, which adopts RL to adjust the driving style of the planner by dynamically modifying the prediction horizon length of the motion planner in real time adaptively w.r.t the event of a change in environment, typically triggered by traffic participants' switch of intents with different driving styles.
The framework models the interaction between the autonomous vehicle and other traffic participants as a Markov Decision Process. 
A temporal sequence of occupancy grid maps are taken as inputs for RL module to embed an implicit intention reasoning. Curriculum learning is employed to enhance the training efficiency and the robustness of the algorithm. We applied our method to narrow lane navigation in both simulation and real world to demonstrate that the proposed method outperforms the common alternative due to its advantage in alleviating the social dilemma problem with proper negotiation skills.

\end{abstract}
%%%%%%%%%%%%%%%%%%%%%%%%%%%%%%%%%%%%%%%%%%%%%%%%%%%%%%%%%%%%%%%%%%%%%%%%%%%%%%%%

% \begin{keywords}

% Cooperative automation, connected vehicles, ADMM, optimal control.

% \end{keywords}

%%%%%%%%%%%%%%%%%%%%%%%%%%%%%%%%%%%%%%%%%%%%%%%%%%%%%%%%%%%%%%%%%%%%%%%%%%%%%%%%
\iffalse
\bibliography{ref/refs}
\fi

\section{Introduction}

% \blue{Note: please use "$\backslash$blue\{content\}" to mark the contents you modified.}

%With the potential to improve traffic efficiency and safety, autonomous driving has received increasing research efforts in the last decades. Normally, the mainstream automatic driving scheme is a hierarchical structure, where the problem is divided into perception, prediction, planning and control~\cite{paden2016survey,ros2012visual}. Taking the output from previous stacks, motion planning module takes the responsibility to find a dynamically feasible, collision free and comfortable resulting path or trajectory, which is essential for the safety of autonomous framework~\cite{katrakazas2015real}.

%General formulation is built as an optimization or search problem, which fulfills the requirements of safe and comfortable by maximizing a complex reward function. However, the real driving scenario is full of dynamic obstacles and complex \blue{interactions} among different traffical participants, where prediction and replanning are required to generate safe trajectories~\cite{schwarting2018planning}. It is necessary to deal with other vehicles' uncertain intentions and complex interacitons.

With promising potential to improve traffic efficiency and safety, the history of Autonomous Vehicles (AVs) could be traced back to the first half of the twentieth century~\cite{kroger2016automated}. Motion planning is one of the critical capabilities for AVs~\cite{katrakazas2015real}. In order to take its responsibility to find a dynamically feasible, collision free, and comfortable resulting trajectory, motion planning module need to identify uncertainties in dynamic environments~\cite{latombe2012robot}. General formulation is built as optimization problems, which fulfill the requirements of safety and comfortability by maximizing a complex reward function.
However, AVs are exposed to complex interactions among traffic participants in real world driving where prediction and re-planning are required to generate safe trajectories~\cite{schwarting2018planning}. It requires the module capable of understanding and adapting to the participants' intents and driving styles by responding in predictable ways without explicit communication.

In order to overcome the challenge, in this paper we propose a negotiation-aware planning framework that integrates reinforcement learning (RL) as a negotiation skill selector into motion planning. We model the interaction between the AV and the other traffic participants as a Markov Decision Process (MDP) under such framework. The key idea is adopting RL to adjust planning modality by dynamically modifying hyper-parameters of the motion planner in real time according to the event of a change in environment, typically triggered by traffic participants' switch of intents with different driving styles. By taking advantages of RL, which tends to have a promising performance in dealing with dynamic environments, the approach embeds implicit prediction and reasoning for the intentions of other vehicles with different driving styles in our framework with a temporal sequence of occupancy grid maps (OGMs) as inputs~\cite{mohajerin2019multi}.
%The sidewalk problem is a benchmark problem for cooperative stochastic games and interactive decision making, in which the agents have to deviate from the optimal pat{}h considering \blue{their} original intersect guidance trajectories~\cite{iselesidewalk}. In the autonomous driving, n{}arrow lane driving is a typical interaction decision making scenario which could be formulated as a sidewalk problem.
%In the sidewalk problem, it is usually hard to reason intentions of interactive agents let alone performing interactively and have an efficient driving procedure. For the reason of biased estimation, interacitve vehicles may perform aggressively and reach to a bottle neck situation or both yield to others and driving tentatively, which lead to the social dilemma problem~\cite{schwarting2019social}.

Narrow lane navigation is one of the typical real world scenarios for autonomous driving, which could be formulated as a sidewalk problem. The sidewalk problem is a benchmark problem for cooperative stochastic games and interactive decision making, in which the agents have to deviate from the optimal path considering their original intersect guidance trajectories~\cite{iselesidewalk}. It is usually hard to reason intentions of other agents let alone interacting to drive efficiently. For the reason of biased estimation, interactive vehicles may perform aggressively and reach to a bottleneck situation or both vehicles yield to others and driving tentatively, which lead to the social dilemma problem~\cite{schwarting2019social}. We applied our method to narrow lane navigation to deal with the sidewalk problem in both simulation and real-world driving to demonstrate that the proposed method outperforms the common alternative due to its advantage in alleviating the social dilemma problem with proper negotiation skills.

Thus the main contributions of this paper are as follows: (1) We introduce an interactive motion planning approach and establish a mapping from the length of planner's prediction horizon to the negotiation skills. (2) Based on the mapping relations, we propose a negotiation-aware planning framework by regulating the prediction horizon via RL to adjust the planning modality adaptively in real time. (3) The framework is implemented in narrow lane navigation and demonstrated the capability for dealing with the sidewalk problem.
The rest of this paper is organized as follows.
Section II reviews related literatures.
Section III proposes our problem statement and negotiation-aware planning framework. Then we look into simulation and experiment results in section IV and conclude this work in section V.

%%%%%%%%%%%%%%%%%%%%%%%%%%%%%%%%%%%%%%%%%%%%%%%%%%%%%%%%%%%%%%%%%%%%%%%%%%%%%%%%%%%%%%%
\iffalse
\bibliography{ref/refs}
\fi

\section{Related Work}

Many works have investigated the methods for motion planning based on graph search, sampling and optimization. A*, RRT, and their variations are classical approaches applied in {autonomous} driving~\cite{bacha2008odin,leonard2008perception}. Considering the structured constraints of road lane, decomposing lateral and longitudinal movement planning in Frenet-frame is a promising method to simplify the optimization based planning in {practice}~\cite{fan2018baidu}. To model the inherent unpredictability of other traffic and the resulting uncertainty, a semi-reactive trajectory generation method based on polynomial approximation is proposed, which can be tightly integrated into the behavioral layer~\cite{Werling2010Optimal}.

Most of the planning schemes are integrated with prediction of other vehicles' future trajectories to avoid collisions~\cite{tran2013modelling}. However, the interaction scenarios in real driving is too complex to be handled perfectly by the prediction module~\cite{dong2017intention}. Besides deducing the intentions of other human traffic participants, driving behavior of {the AV} also need to be {inferred} by other vehicles~\cite{schwarting2018planning}. {Modeling} interactions as multi-agent game~\cite{bahram2015game} or cooperative optimization~\cite{wang2018parallel,lenz2016tactical}, which will generate behavior more interactive rather than reactive. However, both modeling the game and dealing {with} the complexity of solution are all intriguing problems. Partially observable Markov decision process (POMDP) is also a normal way of executing the interaction problem, which assuming that the intentions and re-planning procedures of the other agents are not directly observable and are encoded in hidden variables~\cite{luo2018porca,hubmann2017decision}. In {such situation}, finding a suitable symbolic representation for POMDP is challenging~\cite{somani2013despot}.

Learning-based approach is a promising way for autonomous driving~\cite{zeng2019end,rosbach2019driving}. Reinforcement learning (RL) has shown {its benefits} in many fields, especially in dynamic environment with complex {interactions}~\cite{silver2016mastering,mnih2015human}. By fitting a policy for sequential decision problems, RL has been applied to self-driving~\cite{sallab2017deep}. The framework integrating RL and planning could combine the properties of both algorithms, in which planning is responsible for safety by restricting collision actions and RL is used to reason the intentions of other vehicles and {handle} the interaction uncertainty. Kesleman \emph{et al.} used RL to train a heuristic represented by a deep neural networks and combine it with A*~\cite{keselman2018reinforcement}. Scholz \emph{et al.} proposed a method in which the RL aims to learn the optimization criteria for model-based planning~\cite{scholz2010combining}. For tactical decision making, Hoel \emph{et al.} introduced a general framework, which combines planning and learning in the form of Monte Carlo tree search and RL~\cite{hoel2019combining}.

In this work, we {propose} a framework integrating RL and {planning} which takes a sequential of OGM as inputs of the learning algorithm. Compared with other combined schemes which learn cost map, heuristic search space or subgoals~\cite{rosbach2019driving,keselman2018reinforcement,scholz2010combining,hoel2019combining}, our approach provides a way to adaptively {adjust planning modality referring to negotiation skill selection w.r.t the event of a change in the environment. We utilize the benefits of RL to implicitly reason the intentions of other traffic participants under different driving styles and select negotiation skill respectively while taking advantage of the planner's property to keep safe from collisions in the dynamic environment.
As a result, our framework has the advantage of reasoning behaviors of other traffic participants and adopting an appropriate planning skill to avoid social dilemmas compared to traditional planners which are usually predefined with one fixed driving fashion limited to prior knowledges of algorithm designers.}

%%%%%%%%%%%%%%%%%%%%%%%%%%%%%%%%%%%%%%%%%%%%%%%%%%%%%%%%%%%%%%%%%%%%%%%%%%%%%%%%%%%%%%%
% \input{section/preliminaries}
\iffalse
\bibliography{ref/refs}
\fi

%%%%%%%%%%%%%%%%%%%%%%%%%%%%%%%%%%%%%%%%%%%%%%%%%%%%%%%%%%%%%%%%%%%%%%%%%%%%%%%%%%%%%%%%

\section{Negotiation-aware Adaptive Planning}

% The negotiation-aware planning framework combines model-based planning and RL to deal with autonoumous driving social dilemma caused by interactive decision making. In this section, we formulate the narrowlane driving problem and introduce details of our framework.
% Within our approach, gudiance trajectories are generated by a motion planner embedded with prediction to estimate the intention other of vehicles.
% The key idea is that a policy network is trained to regulate planning behavior adaptively by tunning the prediction horizon online according to the interaction with other vehicles.

The negotiation-aware planning framework integrates planning and RL to alleviate driving social dilemmas caused by interactive driving scenarios. In this section, we formulate the narrow lane navigation problem as an MDP under our proposed framework and introduce details of the framework.

%%%%%%%%%%%%%%%%%%%%%%%%%%%%%%%%%%%%%%%%%%%%%%%%%%%%%%%%%%%%%%%

% \subsection{Markov Decision Process}

% {MDP is a mathematical framework for modeling decision making process from an agent to achieve a goal by learning from a continual interaction with the environment. The agent selecting actions and the environment responding to these actions with rewards and presenting new situations to the agent. A MDP is a tuple $\langle S,A,P,R,\gamma \rangle$, $S$ is a finite set of states represent the environment's situations, $A$ is a finite set of possible actions for the agent, $P$ denotes for the state transition probability function $P=P(S_{t+1}=s'|S_t=s,A_t=a)$, $R$ denotes for the reward function $R=E[R_{t+1}|S_t=s,A_t=a]$, and $\gamma$ is a discount factor that $\gamma\in[0,1]$. Policy $\pi$ describe the agent behavior that is a distribution over actions given states. The goal is to find the optimal policy $\pi^*$ in a MDP by maximizing the cumulative discounted rewards with finite horizon $H$:
% \begin{equation}\label{equ:opt_pro}
% \begin{aligned}
% \bm{\pi^*} = \underset{\bm{\pi}}{\text{argmax}} & \sum_{t=0}^{H}E[\gamma^{t}R(s_t,\pi(s_t))] \\
% \end{aligned}
% \end{equation}
% where the reward function $R$ receives the current state $s_t$ and the action determined by current policy $\pi$ given $s_t$.}

%%%%%%%%%%%%%%%%%%%%%%%%%%%%%%%%%%%%%%%%%%%%%%%%%%%%%%%%%%%%%%%%%%%%%%%%%%%%%%%%%%%%%%%%

\begin{figure}[t]%[hb]
\centering
  \setlength{\abovecaptionskip}{0pt}
  \setlength{\belowcaptionskip}{0em}
\includegraphics[scale=0.2]{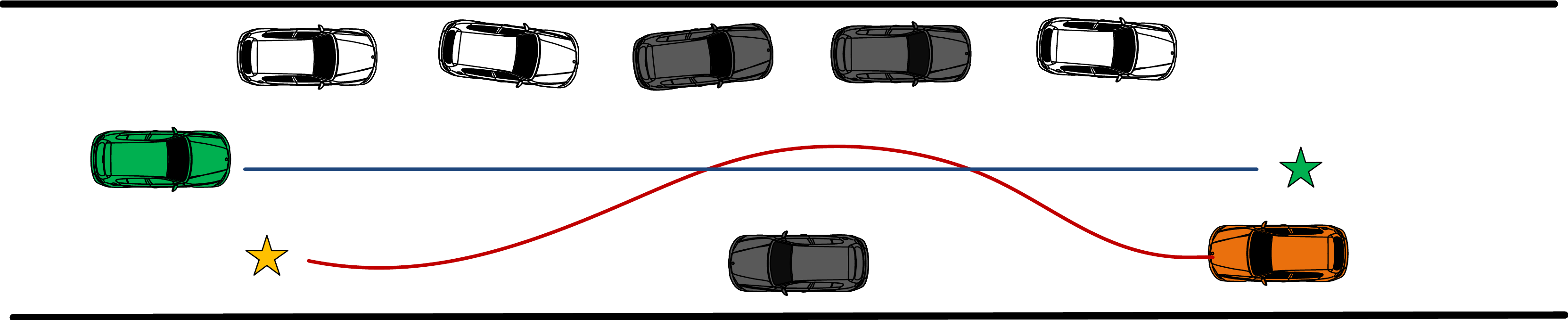}
\caption{The {narrow lane navigation} scenario. Both of the vehicles wish to pass the narrow corridor as soon as possible and keep safe at the same time.}
\label{fig:narrow_driving}
\end{figure}

%%%%%%%%%%%%%%%%%%%%%%%%%%%%%%%%%%%%%%%%%%%%%%%%%%%%%%%%%%%%%%%%%%%%%%%%%%%%%%%%%%%%%%%%

\subsection{Narrow lane navigation}

{We focus on investigating the narrow lane navigation, a typical real world scenario that could be formulated as a benchmark sidewalk problem}. As shown in Fig~\ref{fig:narrow_driving}, there exists a narrow corridor due to {the on street parking, which is common in residential communities}. {The goal for both the AV and the social vehicle} is to pass through the corridor and reach their targets on the other side considering both {safety and efficiency}.
%%%%%%%%%%%%%%%%%%%%%%%%%%%%%%%%%%%%%%%%%%%%%%%%%%%%%%%%%%%%%%%%%%%%%%%%%%%%%%%%%%%%%%%%

\subsection{Negotiation-aware Planning Framework}

\begin{figure*}[t]%[hb]
\centering
  \setlength{\abovecaptionskip}{0pt}
  \setlength{\belowcaptionskip}{0em}
\includegraphics[scale=0.75]{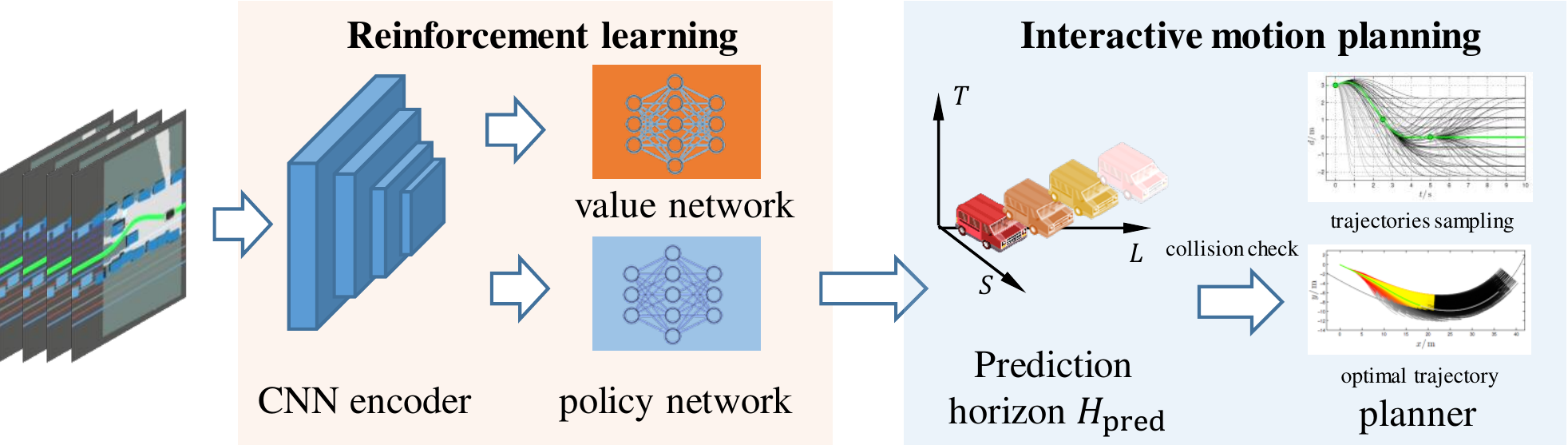}
\caption{An illustration of the relationship between {the} planning behavior and {the} prediction horizon. When considering shorter future trajectories of {social} vehicles, the ego vehicle tends to be more aggressive and vice versa.}
\label{fig:frame_work}
\end{figure*}

%The backbone of our architecture consists of two main parts including the network which outputs prediction horizon regulating policy and an interactive motion planning module responsible for generating feasible trajectories. As shown in Fig \ref{fig:frame_work}, at every time step, a \blue{sequential} OGM which is generated by the perception system from real-world environment will be given into the network then outputs a prediction horizon. Taking into horizon regulating policy as well as guidance route and object list form perception system, the planner will give a future trajectory.

{Our proposed framework consists of two parts including a RL module that outputs planner's prediction horizon and an interactive motion planning module responsible for generating feasible trajectories. As shown in Fig \ref{fig:frame_work}, at every time step, a temporal sequential OGM generated by a perception system that represents the current state of the environment is given to the network, then taking into the selection of prediction horizon from RL module as well as guidance route and objects information from perception system, the planner will give a future trajectory.}

%The RL is composed of a CNN encoder, a value network and a policy network. The CNN encoder is used to process the sequential image information of the OGM images. The value and policy networks, both of which are \blue{fully} connected networks were trained by RL. We applied ACKTR, a policy gradient algorithm to optimize our training procedure~\cite{wu2017scalable}. As for the various of interactive decision making scenarios, we leveraged curriculum learning to handle it~\cite{bengio2009curriculum}.
% The details of training procedure will be dicussed in later subsections.
The RL module is composed of a CNN encoder, a value network and a policy network. The CNN encoder is used to distill the information of history environment dynamics from the sequential OGM images. Both of the value and policy networks are fully connected networks trained to obtain the optimal policy for regulating planner's prediction horizon by achieving the desired goal. We applied ACKTR algorithm, a policy gradient algorithm to optimize our training procedure~\cite{wu2017scalable}. As for the various of interactive situations, we leveraged curriculum learning to handle it~\cite{bengio2009curriculum}.

%%%%%%%%%%%%%%%%%%%%%%%%%%%%%%%%%%%%%%%%%%%%%%%%%%%%%%%%%%%%%%%%%%%%%%%%%%%%%%%%%%%%%%%%

\subsection{Interactive Motion Planning}

Motion planning could be formulated as an optimization problem as follows:
\begin{equation}\label{equ:opt_pro}
\begin{aligned}
 \underset{\bm{x}}{\text{argmin}} \quad & \sum_{t\in[0, H_{\rm plan}]}c(x_t)\\
 \text{subject to} \quad & f_t(x_t, \dot{x}_t, ...) \leq 0 \quad t\in[0, H_{\rm plan}],\\
\end{aligned}
\end{equation}
where $t$ is time step, $H_{\rm plan}$ is the planning horizon, $\bm{x}=\{x_1, x_2, ..., x_{H_{\rm plan}}\}$ represents the planning trajectory, $c$ and $f_t$ are cost function and constraints considering safety, vehicle dynamics and comfort respectively. Since finding an optimal solution of the problem \eqref{equ:opt_pro} is PSPACE-hard with the nonholonomic constraints, we approximate the solution by sampling a set of physically valid trajectories and pick the one with minimum cost.

We adopt a semi-reactive motion planner by means of optimal-control strategies~\cite{Werling2010Optimal}. The planner takes inputs include HD map and LiDAR sweeps for navigation guidance reference and objects array collision checking. The trajectory set is generated by polynomial approximation that quintic polynomials are the jerk-optimal connection between two secondary order states $[x_0, \dot{x}_0, \ddot{x}_0]$  and $[x_{H_{\rm plan}}, \dot{x}_{H_{\rm plan}}, \ddot{x}_{H_{\rm plan}}]$. Lateral and longitudinal movements are generated by sampling along Frenet-Frame independently and checked against acceleration values of dynamic limitation before combination. The trajectory with the lowest conjoint cost function and free of collision is picked out for the downstream controller to track.

\begin{figure}[t]%[hb]
\centering
  \setlength{\abovecaptionskip}{0pt}
  \setlength{\belowcaptionskip}{0em}
\includegraphics[scale=0.42]{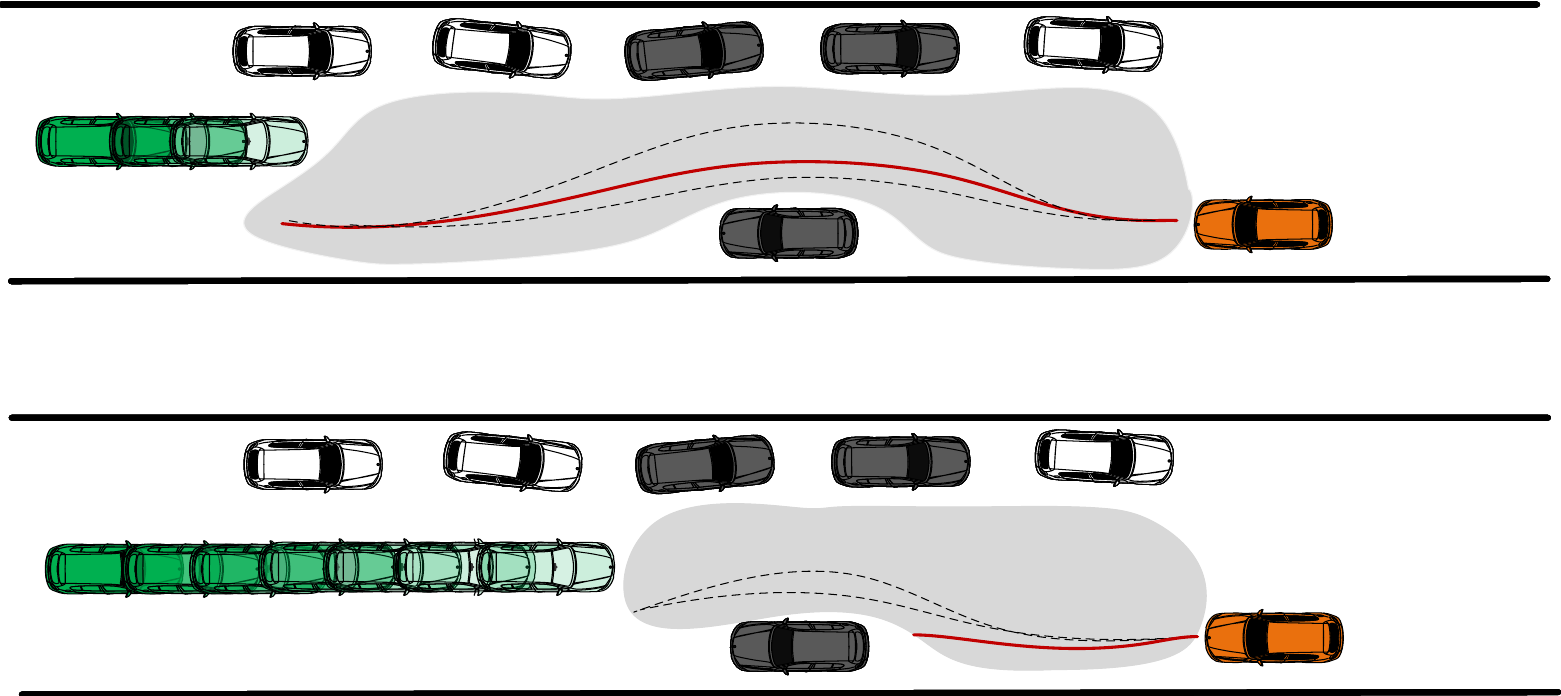}
\caption{An illustration of the relationship between planning behavior and prediction horizon. When considering shorter future trajectories of interactive vehicles, the ego vehicle tends to be more aggressive and vice versa.}
\label{fig:beh_n_hor}
\end{figure}

In order to consider the interaction with other social vehicles, a prediction module is embedded into our motion planner. We use $\bm{\hat{x}} = \{ \hat{x}_1, \hat{x}_2, ..., \hat{x}_{H_{\rm pred}} \}$ to denote the $H_{\rm pred}$ steps rollout for a prediction trajectory of each vehicle. The prediction trajectory is projected onto the SLT frame for the collision checking. As default, the length of prediction horizon $H_{\rm pred}$ should be the same as planning horizon $H_{\rm plan}$. The prediction trajectories of other vehicles are hardly to be precise, longer prediction horizon normally lead to a relatively conservative choice of plans and shorter ones could otherwise lead to a more aggressive driving behavior as shown in Fig~\ref{fig:beh_n_hor}.

AVs requires conservative behaviors for safe operations because of lacking capability of precise predictions for dynamics. However, social vehicles with various driving styles could react to our AV differently, thus being conservative constantly could have a passive influence on driving efficiency or even lead to traffic bottlenecks, especially when the other vehicles also drive in a modest fashion. On the other hand, being blindly aggressive without proper negotiations could still cause traffic dilemmas.

%%%%%%%%%%%%%%%%%%%%%%%%%%%%%%%%%%%%%%%%%%%%%%%%%%%%%%%%%%%%%%%%%%%%%%%%%%%%%%%%%%%%%%%

\subsection{RL formulation}

\subsubsection*{\textbullet \quad Markov Decision Process}

MDP is an mathematical framework for modeling decision making process for an agent to achieve a goal by learning from a continual interaction with the environment. The agent selects actions and the environment responds to these actions with rewards and presents new situations to the agent. An MDP is a tuple $\langle S,A,P,R,\gamma \rangle$, where $S$ is a finite set of states represent the environment's situations, $A$ is a finite set of possible actions for the agent, $P$ denotes the state transition probability function $P\!=\!P(S_{t+1}\!=\!s'|S_t\!=\!s,A_t\!=\!a)$, $R$ denotes for the reward function $R\!=\!E[R_{t+1}|S_t\!=\!s,A_t\!=\!a]$, and $\gamma$ is a discount factor that $\gamma\in[0,1]$. Policy $\pi$ describes the agent behavior that is a distribution over actions given states. The goal is to find the optimal policy $\pi^*$ in an MDP by maximizing the cumulative discounted rewards with finite horizon $H$:
\begin{equation}\label{equ:opt_pro}
\begin{aligned}
\bm{\pi^*} = \underset{\bm{\pi}}{\text{argmax}} & \, E[\sum_{t=0}^{H} \gamma^{t}R(s_t,\pi(s_t))] \\
\end{aligned}
\end{equation}
where the reward function $R$ receives the current state $s_t$ and the action determined by current policy $\pi$ given $s_t$.

\subsubsection*{\textbullet \quad State and action {representation}}

%The sequential temporal occupancy grid map is used as the state representation of RL. An OGM is a map of size $W \times H$ pixels rendered into the top-down coordinate system, which represent the occupancy state of regions around the ego-vehicle. The sequential OGM are sampled according to a fixed time step $\delta t$ perception information from past $T_{\rm ogm}$ seconds. Rather than using the raw history OGM map directly, coordination transform based on current vehicle-link frame is implemented. This makes the sequential representation could show all the potential dynamic objects from current view.

%The OGM map is easy to generate in the simulation environment or could be constructed by the meta-data form LiDAR in real-world driving.
%The sequential representation of the dynamic environments is inspired from ChuffeeurNet. Compared with the raw \blue{sensor} data representation, OGM shows its \blue{advantages} in flexibility and generalization\cite{BansalChauffeurNet}.

An OGM is a grid map of size $L_{\rm W} \times L_{\rm H}$ pixels rendered into the top-down coordinate system, which represent the occupancy state of regions around the ego-vehicle. It is easy to generate in the simulation environment or could be constructed by the meta data form LiDAR in real world driving. Compared with the raw sensor data representation, OGM shows its advantages in flexibility and generalization\cite{BansalChauffeurNet}.

As the state representation of RL, a sequence of temporal OGMs are sampled according to a fixed time step $\Delta t$ from past $T_{\rm ogm}$ seconds, thus the inputs dimension is $[T_{\rm ogm} \times L_{\rm W} \times L_{\rm H}]$. Coordination transformation based on current vehicle-link frame is implemented in order to make the temporal sequential representation could indicate the potential dynamics from current view.

%For action space, we choose the discrete prediction horizon in the dimension of $T_{\rm pre}$ for the autonomous vehicle.

According to the mapping relations from planner's prediction horizon to planning modality, we discretized the prediction horizon into a $H_{\rm pred}$ dimension vector which refers to different driving styles from aggressive to conservative.

\subsubsection*{\textbullet \quad Reward setting}

%The task is designed to \blue{terminate} when all vehicles get their target position or collision happens during the episodic. We consider safety, efficiency and task completion to design the reward function. To improve efficiency, a negative step reward is given every time step. To enhance safety, a penalty is given to collision case. To encourage task completion, there is a positive reward when vehicles finish their driving task. The reward setting are shown in \eqref{equ:reward}.

{The task is designed to terminate and reset either all vehicles reach their target positions or collision. Since the step by step interaction process between the AV and the social vehicle is hard to be intuitively described, we consider the safety, efficiency and task completion for reward function design in a sparse manner. A negative reward is given to penalize collision and a positive reward for the vehicles both finish their driving tasks. Also, a negative per-step reward is accumulatively given with fixed episodic horizon $H$ to encourage efficiency. A detailed reward setting is shown in \eqref{equ:reward}.}

%To improve efficiency, a negative step reward is given every time step. To enhance safety, a penalty is given to collision case. To encourage task completion, there is a positive reward when vehicles finish their driving task. The reward setting are shown in \eqref{equ:reward}.}
% \renewcommand{\thetable}{\arabic{table}}
% \begin{table}[t]
% \centering
% \renewcommand\arraystretch{1.1}
% \caption{Reward setting}
% \begin{tabular} {cc}
% \toprule[1pt]
%  Items & Reward \\
% \hline
% step reward & -0.25  \\
% collision penalty & -50  \\
% task completion & 10  \\
% \bottomrule[1pt]
% \end{tabular}
% \label{tab:reward}
% \end{table}
{
\begin{equation}\label{equ:reward}
R =
\begin{cases}
-0.25 * t,& \rm \textit{per-step penalty,} \\
-5,& \rm \textit{collision penalty,} \\
10,& \rm \textit{reached targets}
\end{cases}
\end{equation}
}

\subsubsection*{\textbullet \quad Algorithm}

%The hierarchical framework where actions of RL do not interact with environment directly will increase the difficulty of training. To enhance the traning efficiency of training, we applied curriculum learning into our training procedure and used ACKTR to optimize the network model.

%ACKTR is a policy gradient method with the trust region optimization method use second order optimizer. By applying the Kronecker-factored approximation to optimize both actor and critic, ACKTR will speed up the optimization and stabilize the training procedure.

The RL algorithm we adopt in this work named actor critic using Kronecker-factored trust region (ACKTR)~\cite{wu2017scalable}. %~\cite{https://arxiv.org/abs/1708.05144}
ACKTR is a policy gradient method with the trust region optimization method use second order optimizer. By applying the Kronecker-factored approximation to optimize both actor and critic, ACKTR will speed up the optimization and stabilize the training procedure.

With an actor critic architecture, we need to optimize a stochastic policy $\pi_{\theta}:S\times A\rightarrow\mathbb{R}_+$ that parameterized by $\theta$. Given the state $s_t$ from $S$, which are assumed to be the OGM based observations of the environment, with actions $a_t$ from the action space $A$ chosen by the policy distribution $\pi_{\theta}(\cdot|s_t)$. The AV interact with the environment with prediction horizon assigned motion planner and perceived next state $s_{t+1}\sim P(s_{t+1}|s_t,a_t)$ and the immediate reward $R(s_t,a_t)\in\mathbb{R}$ according to the reward function defined in \eqref{equ:reward}. Altogether, it can be formalized as an $\gamma$-discounted MDP with the expected discounted cumulative return is:
\begin{equation}\label{equ:policy gradient}
\begin{aligned}
J(\Theta)= E_{\pi}[\sum_{t=0}^{H} \gamma^{t}R(s_t,\pi(s_t))]. \\
\end{aligned}
\end{equation}
Update the $\theta$ w.r.t the policy gradient defined as:
\begin{equation}\label{equ:policy gradient}
\begin{aligned}
\nabla_{\Theta}J(\Theta) = E_{\pi}[\sum_{t=0}^{H} \Psi^{t}\nabla_{\Theta}log\pi_{\theta}(a_t|s_t)], \\
\end{aligned}
\end{equation}
where $\Psi^t$ denotes for the advantage function $A^{\pi}(s_t,a_t)$. The advantage is calculated as follows:
\begin{equation}\label{equ:policy gradient}
\begin{aligned}
%A^{\pi}(s_t,a_t)=(R(s_{t},a_{t})+V^{\pi}_{\phi}(s_{t+1}))-V^{\pi}_{\phi}(s_{t}), \\
A^{\pi}(s_t,a_t)=\sum_{i=0}^{k-1} (\gamma^{i}R(s_{t+i},a_{t+i}) + \gamma^{k}V^{\pi}_{\phi}(s_{t+k}))-V^{\pi}_{\phi}(s_{t}),\\
\end{aligned}
\end{equation}
where $V^{\pi}_{\phi}(s_{t})$ denotes for the value network that estimate the expected reward with policy $\pi$. We train the network by performing temporal difference updates to minimize
the squared difference between the $k$-step bootstrapped returns $\hat{R}_t$ and $V^{\pi}_{\phi}(s_t)$.

\subsubsection*{\textbullet \quad Curriculum Learning}

In order to enhance the training efficiency and robustness of our algorithm, we employ curriculum learning that could help intelligent agents to learn better with organized examples which gradually illustrate more and more complex concepts.

In our training procedure, the initial position of the social vehicle is selected from a certain range with a driving style assigned. We deploy interactive motion planning method to control social vehicles and define the driving styles from aggressive to conservative w.r.t the mapping from prediction horizons to planning behaviors. Different distributions of both the initial position and the driving style significantly affect the difficulty of our task and the curriculum is designed by decomposing the learning procedure into multiple phases accordingly.

For instance, in first phase we set initial positions far away from the narrow corridor for conservative social vehicles and nearby positions for aggressive ones, than in order to increase the difficulty, the distributions of initial positions will be set closer to each other. Finally, our AV learns to reasoning about the social vehicle's intent according to its driving style and road geometry, and thus react with proper choice of skill w.r.t task situation and goal.
The corresponding training procedure is shown in Algorithm~\ref{alg:Framwork}.
%Because of the boundaries of different driving behavior have obvious discrimination, the curriculum learning starts form a distribution $D_0(T)$ which tends to sample mainly form boundries. The learning procedure is divided into three stage and distribution of task approaches the uniform distribution gradually.
% The process of distribution variation could be illustrated in \ref{fig:curriculum}.

% \begin{figure}[t]%[hb]
% \centering
%   \setlength{\abovecaptionskip}{0pt}
%   \setlength{\belowcaptionskip}{0em}
% % \includegraphics[scale=0.42]{figure/beh_n_hor.pdf}
% \caption{An illustration of the distribution change}
% \label{fig:curriculum}
% \end{figure}

 \begin{algorithm}[t]
  \caption{ Curriculum learning procedure.}
  \label{alg:Framwork}
  \begin{algorithmic}[1]
    \Require
      \begin{varwidth}[t]{\linewidth}
      distribution $D_0$,..., $D_s$, iteration numbers $K_0$,..., $K_s$,
      \\task batch size $M$, number of trajectories $N$
      \end{varwidth}
    \Ensure policy and value networks $\pi_{\theta}$, $V_{\phi}$
    \State Randomly initialize $\theta$, $\phi$
    \While {\emph{not done}}
      \For{$(K, D) \in \{ (K_0, D_0), ..., (K_s, D_s) \}$}
         \For{$k \in \{(1, ..., K)\}$}
            \parState{Sample M number of tasks $T \in D$}
            \For{\textbf{all} $T$}
               \parState{Sample $N$ trajectories $\tau^{1:N}$ with $\pi_{\theta}$}
            \EndFor
            \State update network $\pi_{\theta}$ and $V_{\phi}$ using ACKTR
         \EndFor
      \EndFor
    \EndWhile
  \end{algorithmic}
\end{algorithm}
%

%%%%%%%%%%%%%%%%%%%%%%%%%%%%%%%%%%%%%%%%%%%%%%%%%%%%%%%%%%%%%%%%%%%%%%%%%%%%%%%%%%%%%%%

\iffalse
\bibliography{ref/refs}
\fi

%%%%%%%%%%%%%%%%%%%%%%%%%%%%%%%%%%%%%%%%%%%%%%%%%%%%%%%%%%%%%%%%%%%%%%%%%%%%%%%%%%%%%%%%

\section{Experiment}

\begin{figure*}[ht]
\centering
\subfigure[driving time]{
\label{subfig:push_driving_time}
\begin{minipage}[t]{0.33\linewidth}
\centering
\includegraphics[width=2.5in]{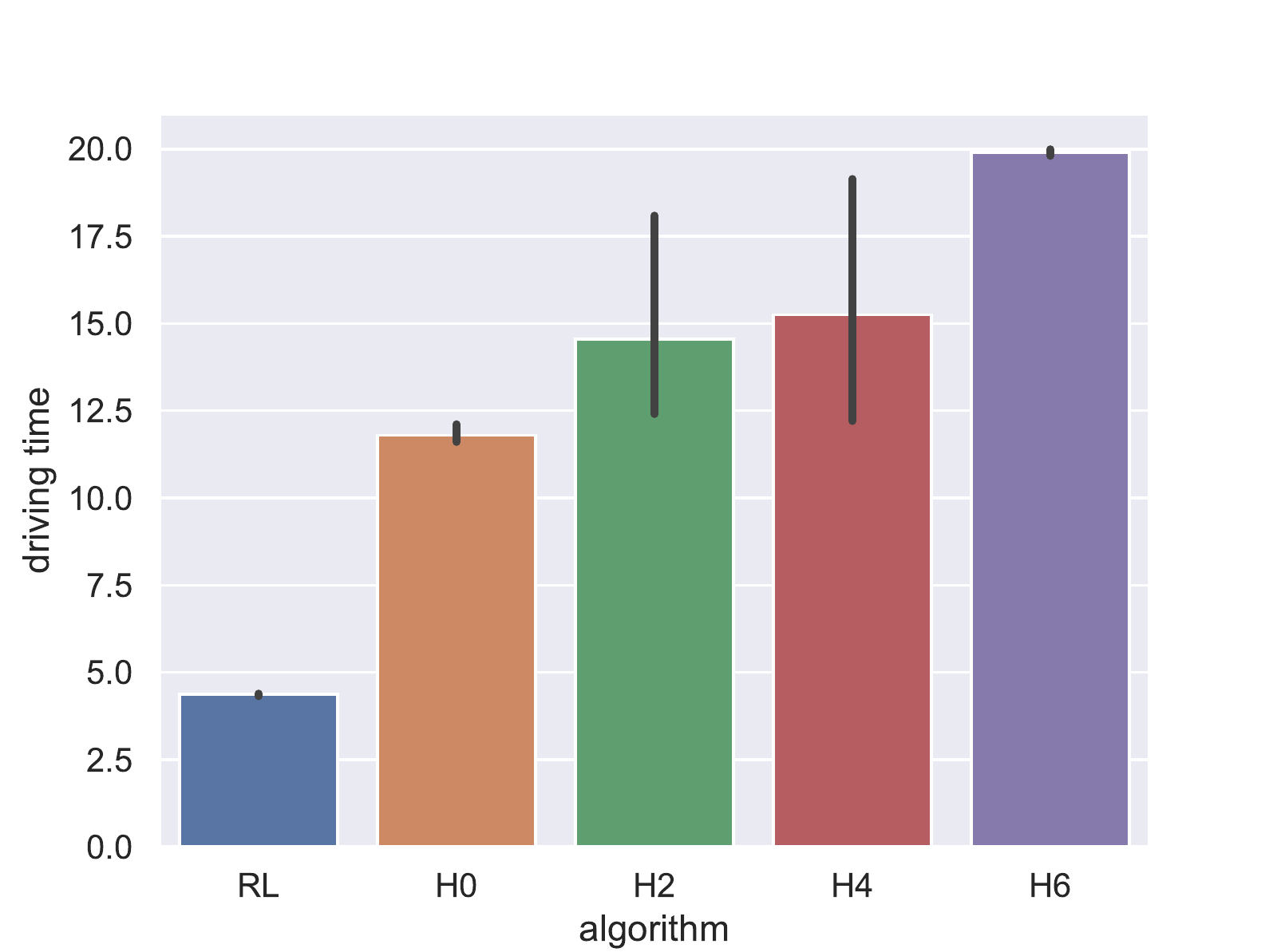}
%\caption{fig1}
\end{minipage}%
}%
\subfigure[suspend number]{
\label{subfig:push_pause_number}
\begin{minipage}[t]{0.33\linewidth}
\centering
\includegraphics[width=2.5in]{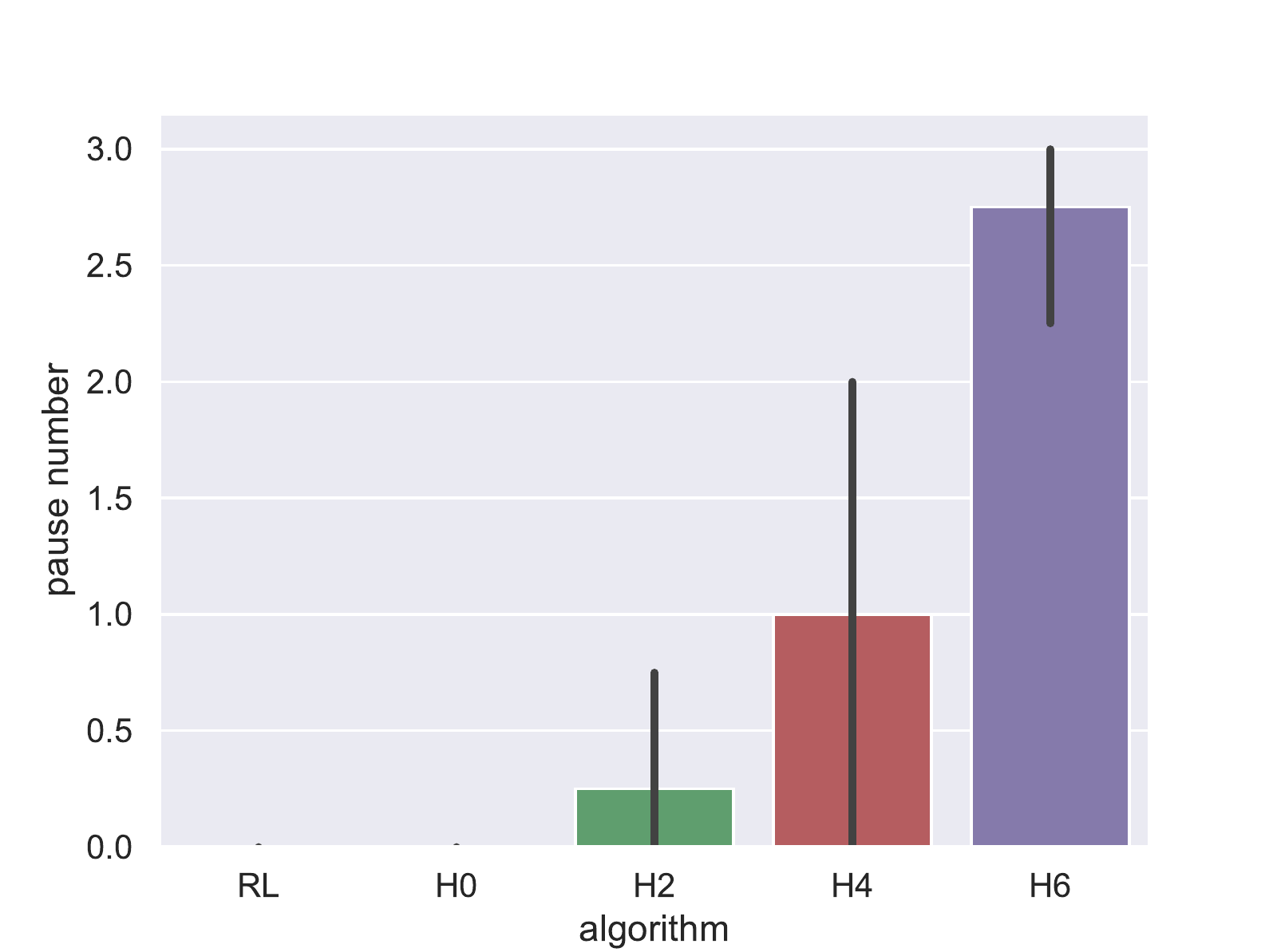}
%\caption{fig2}
\end{minipage}
}%
\subfigure[reward]{
\label{subfig:push_reward}
\begin{minipage}[t]{0.33\linewidth}
\centering
\includegraphics[width=2.5in]{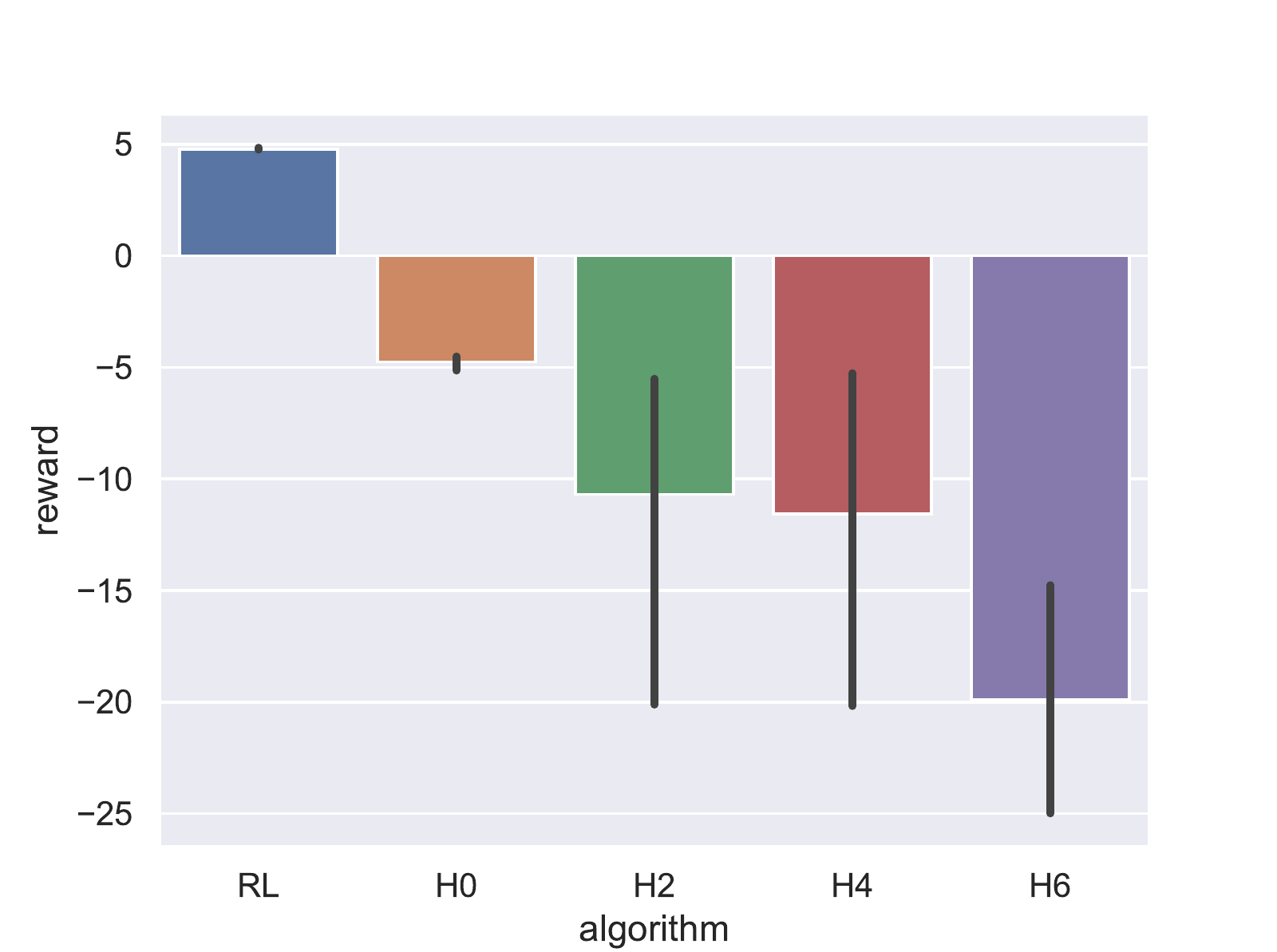}
%\caption{fig2}
\end{minipage}%
}%
\centering
\caption{Performance comparisons of our framework in push experiment setting. As \ref{subfig:push_driving_time} \ref{subfig:push_pause_number} show, all the algorithm could complete driving task in limit time, while the adaptive behavior planner has better driving efficiency and take less suspend time; Thus the proposed framework tends to have a higher reward in \ref{subfig:push_reward}.}
\label{fig:perf_push}
\end{figure*}

\begin{figure*}[ht]
\centering
\subfigure[driving time]{
\label{subfig:yield_driving_time}
\begin{minipage}[t]{0.33\linewidth}
\centering
\includegraphics[width=2.5in]{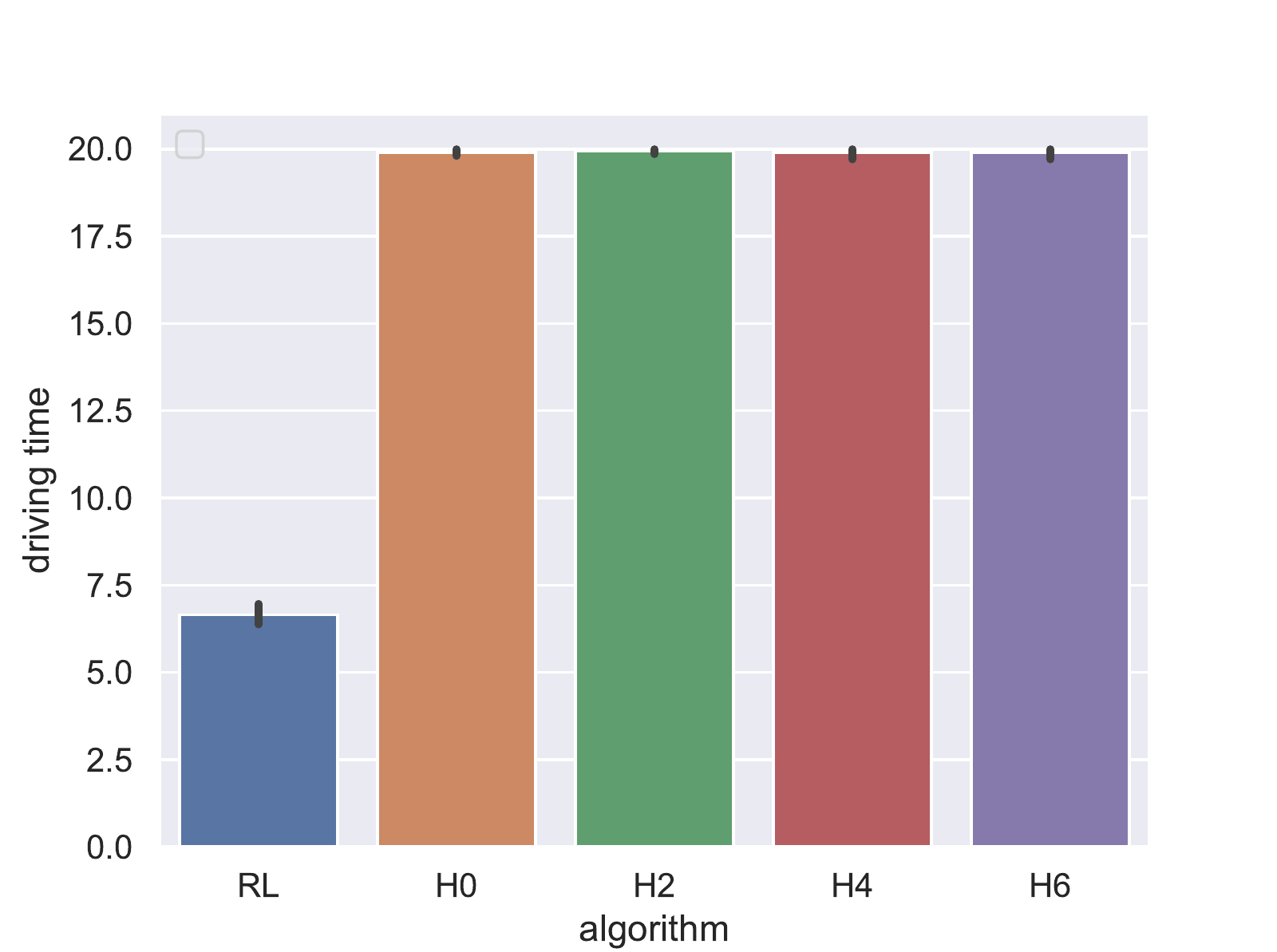}
%\caption{fig1}
\end{minipage}%
}%
\subfigure[suspend number]{
\label{subfig:yield_pause_number}
\begin{minipage}[t]{0.33\linewidth}
\centering
\includegraphics[width=2.5in]{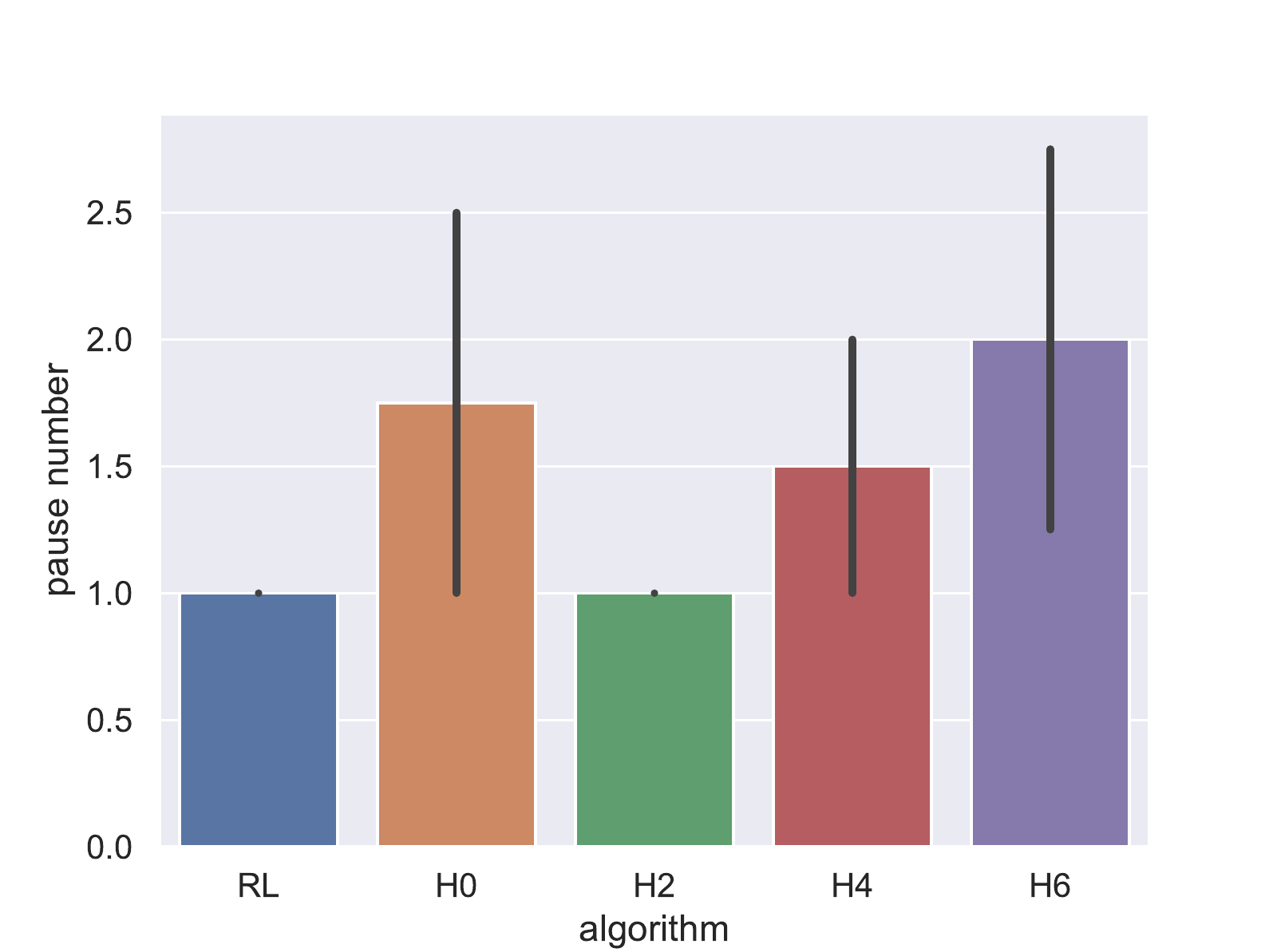}
%\caption{fig2}
\end{minipage}
}%
\subfigure[reward]{
\label{subfig:yield_reward}
\begin{minipage}[t]{0.33\linewidth}
\centering
\includegraphics[width=2.5in]{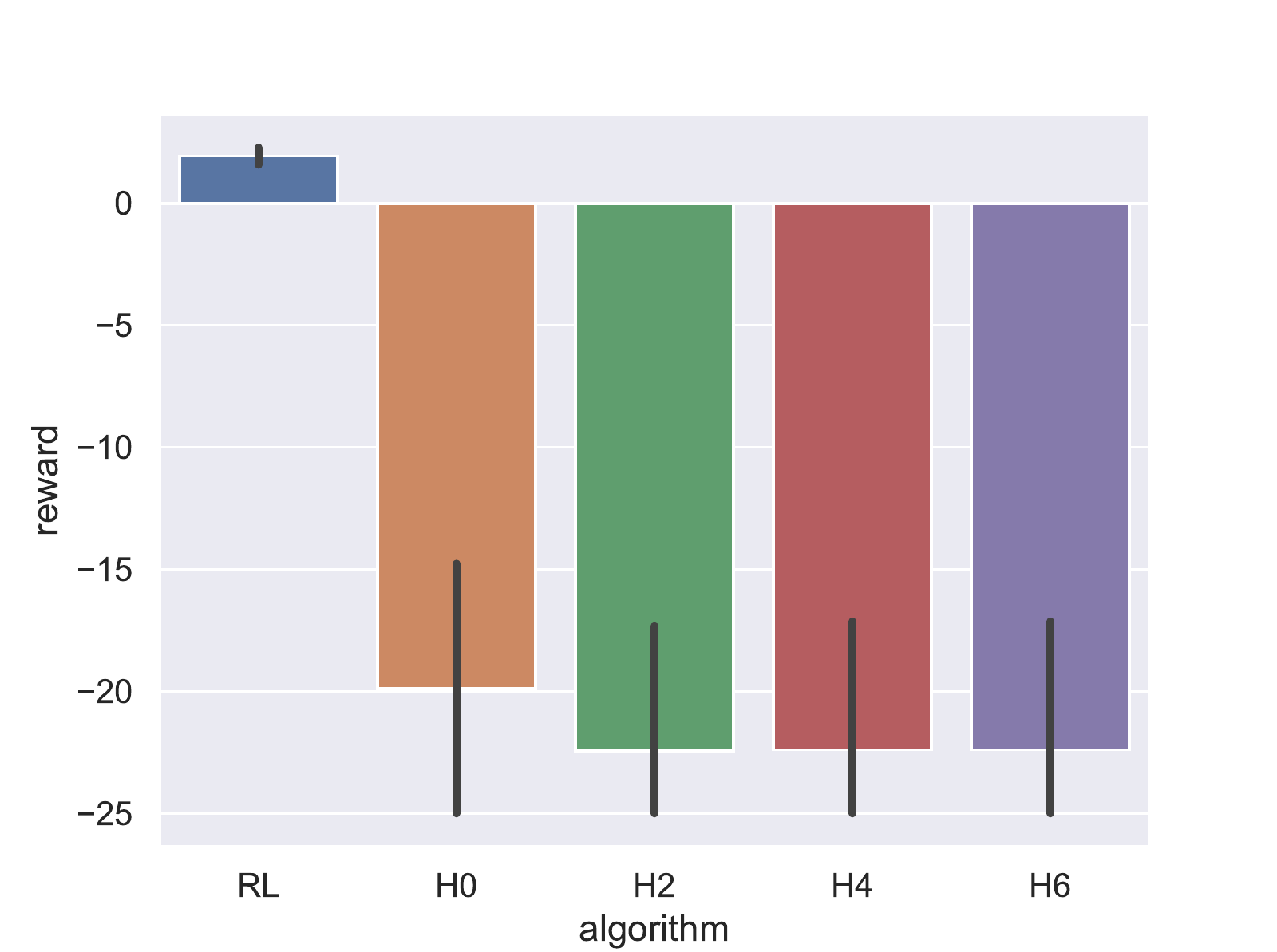}
%\caption{fig2}
\end{minipage}%
}%
\centering
\caption{Performance comparisons of our framework in yield experiment setting. Because of a closer social vehicle distance and the social vehicle's intent of pushing, ego vehicle will yield to keep safe. \ref{subfig:yield_driving_time} shows that except the proposed algorithm, other schemes could not reach the goal position in limit time. To yield to social vehicles, the suspend time tends to be more than that in push experiment setting as shown in \ref{subfig:yield_pause_number}.}
\label{fig:pef_yield}
\end{figure*}

\subsection{Experimental settings}

The proposed algorithm was trained and tested in the narrow lane navigation scenario.
Relative distances from the narrow corridor and driving styles of the two vehicles are the key features to impact the driving behavior. The initial position of the social vehicle is randomly sampled in a range. As for the training and testing in the simulator, a polynomial planner introduced in interactive motion planning section is arranged for the social vehicle which is identical with the one used in our framework. We also implemented and evaluated our work on a real autonomous vehicle in a campus scenario.

\subsection{Implementation details}

The initial positions of the social vehicle are sampled from $11.99$m to $14.02$m for conservative behavior and $16.33$m to $18.80$m for aggressive behavior. As for the ego autonomous vehicle, the position is fixed with $13.35$m.
The frequency of the implementation is 5Hz and we sampled sequential OGM of $84\times84$ size every second in the last 4 seconds, which means that 4 channels of OGM are given to our CNN encoder at a time.
The CNN encoder which extracts features from OGM sequence has 3 convolution layers with the filter size of $32\times8\times8$, $64\times4\times4$ and $64\times3\times3$, followed with 1 fully connected layer with 512 output neurons.
Both value and policy networks are fully connected networks with 2 hidden layers with 128 units. The Policy network has 4 discrete outputs units to represent time horizon $\{ 0, 2, 4, 6 \}$ while the value network has only one output. We use Adam as the optimizer and a learning rate of $10^{-4}$ and $10^{-3}$ for policy network and value networks respectively. For the setting of RL, we use mini-batch size of $m=512$, trajectory length of $H=128$, number of trajectories to collect for each iteration of $N=200$, and $\gamma=0.99$ as the discount factor.

\subsection{Results and discussion}

We tested the negotiation-aware framework in both push and yield experiment settings in narrow lane navigation and demonstrate the effectiveness of our proposed framework compared to the polynomial planner with different fixed prediction horizons in this section.

Fig~\ref{fig:perf_push} and Fig~\ref{fig:pef_yield} illustrate the performance comparisons in the two experiment settings respectively. In the push setting where the social vehicle is further from narrow corridor than ego vehicle, most of the algorithms could dominate autonomous vehicle reach its target within time limit while our proposed algorithm takes less time and suspend times to reach the target. In the yield setting scenario, the vehicles with polynomial planner tends to get into a social dilemma with the almost identical relative distances. On the contrary, our algorithm could decide to yield in advance before stuck in a dilemma by reasoning the other vehicle's behavior.

Generally, compared with the polynomial planner with fixed prediction horizon, our adaptive framework could complete the driving task in shorter time. One reason could explain the outstanding performance in driving efficiency is the less hesitate in driving interaction as shown in Fig~\ref{subfig:push_pause_number} and Fig~\ref{subfig:yield_pause_number}. Another reason is that our framework learns the approximating optimum yield position from the sequential OGM images which describes the environment dynamics. This makes the autonomous vehicle has less chance to get into dilemma situations with social vehicles. As a result, our framework has a higher reward than the polynomial planner with fixed prediction horizon. The statistical results also show that the performance of our proposed framework has better stability because of the robustness w.r.t to environment dynamics consists of social participants and road geometry.
\begin{figure}[t]
\centering
\subfigure[Behavior index in the push experiment setting.]{
\label{subfig:svo_push}
\includegraphics[width=2.3in]{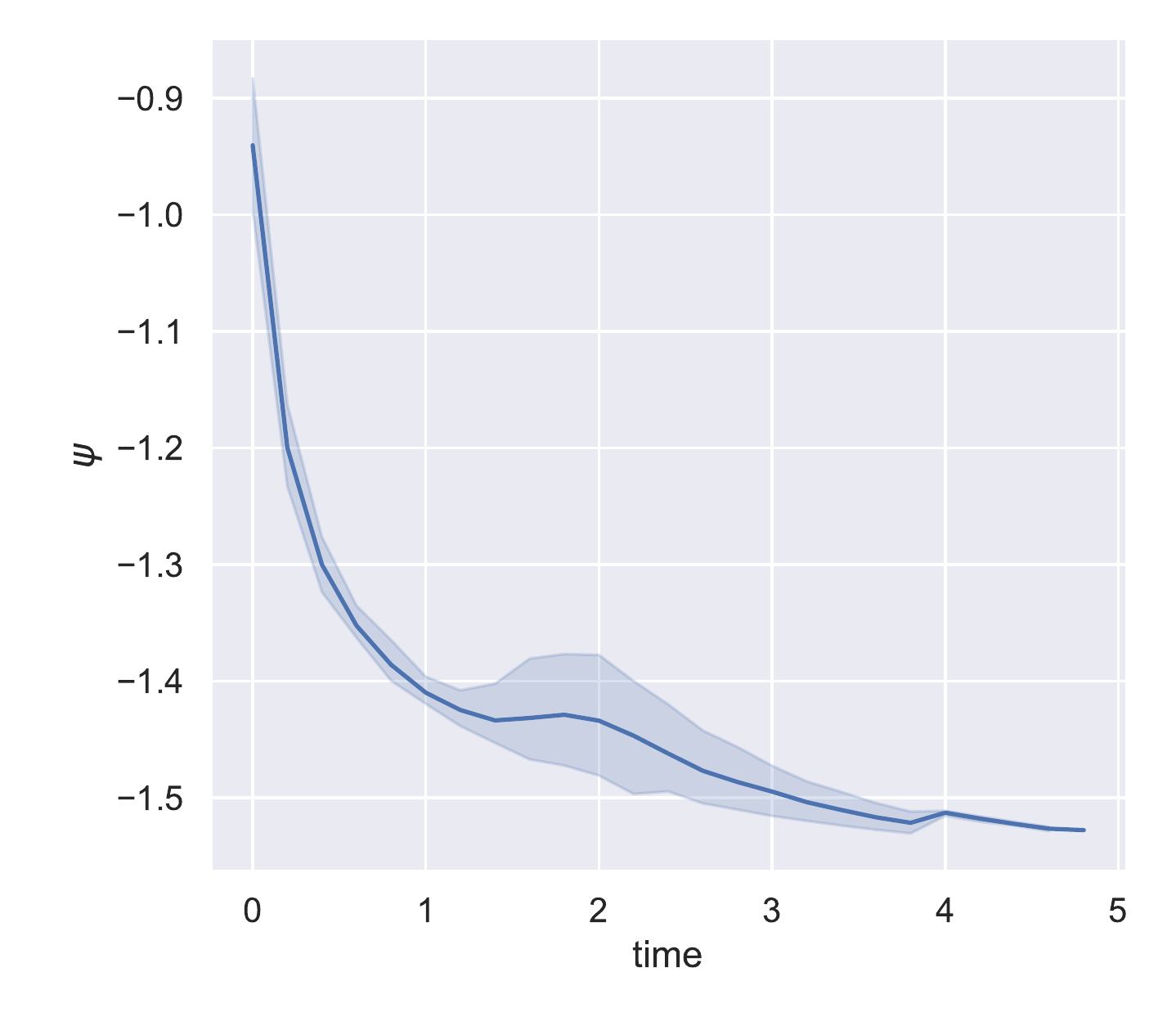}
} \\
\subfigure[Behavior index in the yield experiment setting]{
\label{subfig:svo_yield}
\includegraphics[width=2.3in]{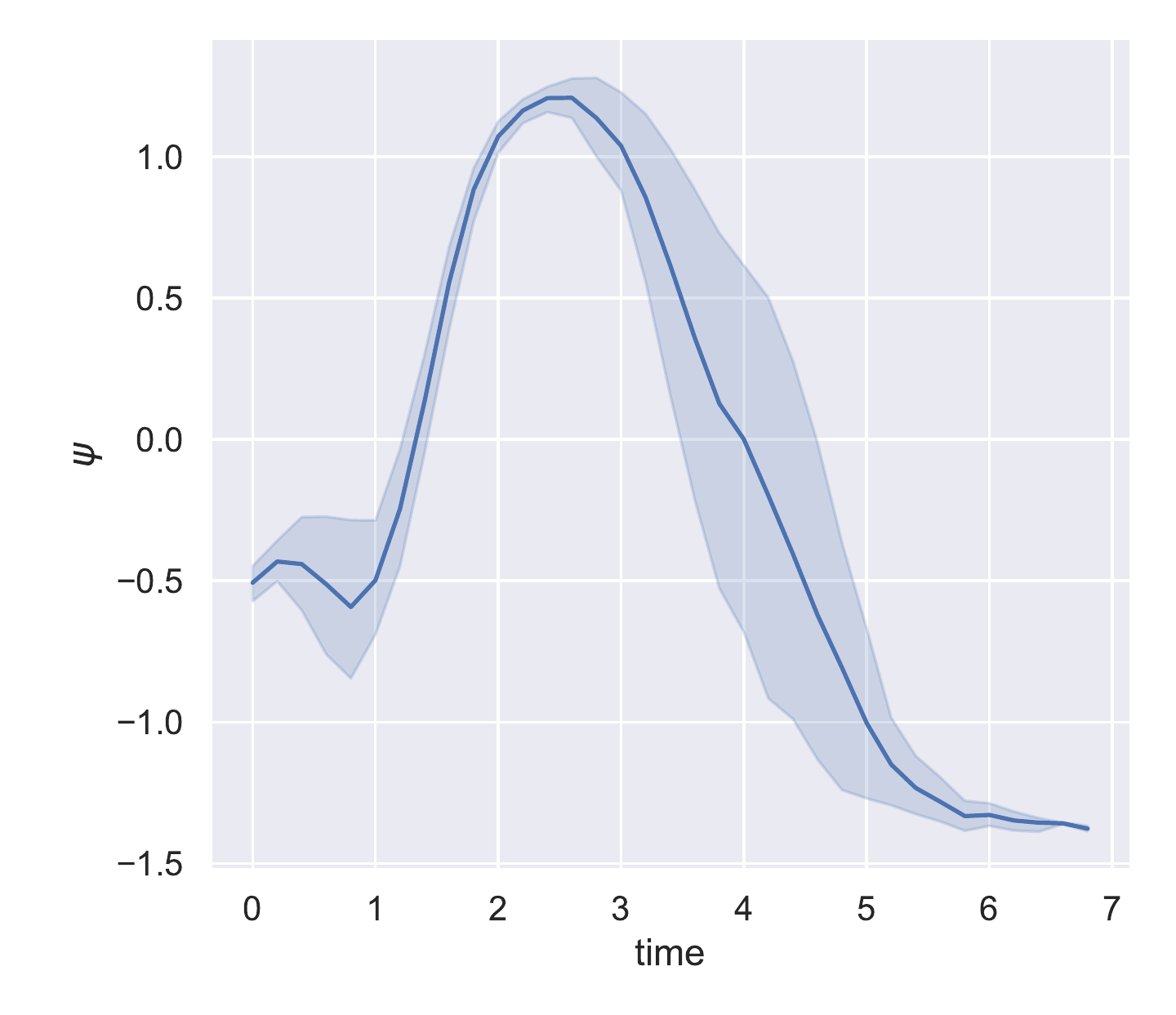}
}
\caption{The change of driving behavior index in narrow lane navigation scenario. In the push experiment setting, the index value is always negative. However in the yield experiment setting, the index value shows that the autonomous vehicle will yield to the social vehicle at first.}
\label{fig:svo_change}
%\vspace{-2mm}
\end{figure}

The behavior of prosocial or egoistic could be measured by Social Value Orientation (SVO), which is an index represents the weight assignment for the interests of ego vehicle and others in the formation of angles~\cite{schwarting2019social}.
Inspired by the work in~\cite{schwarting2019social}, we also use an angle like index as the measurement for the social behavior in sidewalk scenario,
\begin{equation}\label{equ:svo_like}
\begin{aligned}
 \psi = \text{atan}{\frac{s_{\rm soc}-s_{\rm ego}}{d_{\rm ini, ego}/d_{\rm ini, soc}}},
\end{aligned}
\end{equation}
where $d$ is the initial distance from the corridor, $s$ is the travel distance of ego and social vehicles respectively. Both distances are counted at the time step since the social vehicle enter the ego vehicle's perception. Such index works the same as SVO, that the behavior tends to be prosocial for positive values and competitive for a negative ones.
The numerator of \eqref{equ:svo_like} describes the fact that which vehicle dominants the driving interaction and the denominator considers the impact of the initial position.

Fig~\ref{fig:svo_change} shows the change of driving behavior index $\phi$ in push and yield experiment setting respectively. The behavior of autonomous vehicle in the interactive driving and the driving behavior index could verify each other.
In the push scenario the adaptive framework will regulate the prediction horizon to performance aggressively to pass the narrow corridor as soon as possible when the social vehicle is further from the corridor. Fig~\ref{subfig:svo_push} shows negative index values during driving which reflects the aggressive driving behavior.
And in the yield experiment setting. To keep safe and retain driving efficiency at the same time, ego vehicle will pause at a reasonable position to yield to the social vehicle. After the interactive vehicle passes the corridor, our ego vehicle will perform aggressively to reach to the goal. Fig~\ref{subfig:svo_yield} illustrates the driving behavior tunning procedure in this scenario, that the index value is positive and then negative.

\begin{figure}[t]%[hb]
\centering
  \setlength{\abovecaptionskip}{0pt}
  \setlength{\belowcaptionskip}{0em}
\includegraphics[scale=0.75]{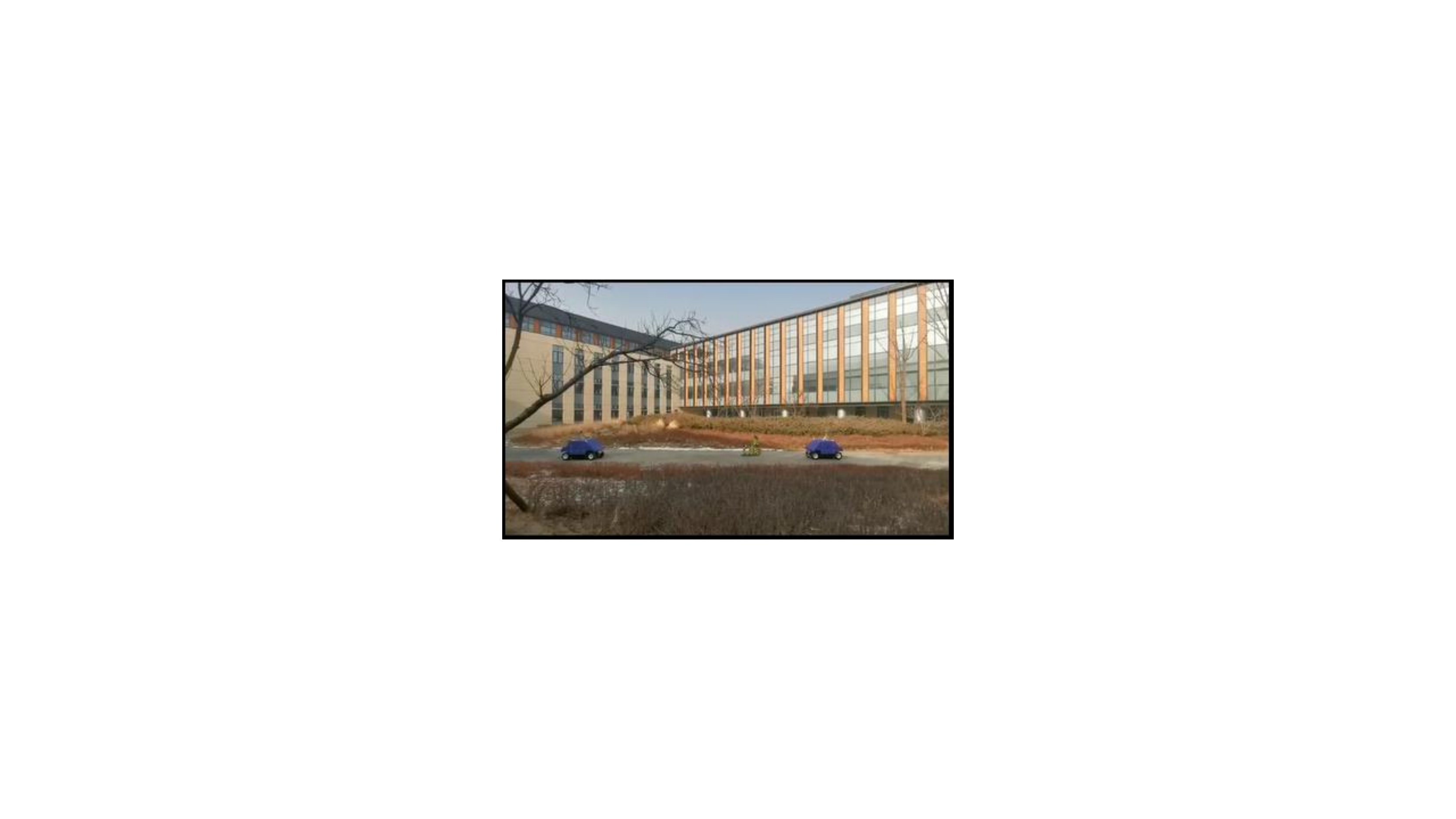}
\caption{Testing of negotiation-aware framework in real-world driving.}
\label{fig:real_world}
\end{figure}

We implemented the framework onto a real autonomous vehicle and tested in a real world narrow lane navigation scenario as shown in Fig~\ref{fig:real_world}. Our proposed approach demonstrates a good generalization in real world driving. A supplementary video is provided to show the performance of our work\footnote{See videos at \url{https://youtu.be/j9FK_RNbbI0}}.

%%%%%%%%%%%%%%%%%%%%%%%%%%%%%%%%%%%%%%%%%%%%%%%%%%%%%%%%%%%%%%%%%%%%%%%%%%%%%%%%%%%%%%%
% \balance
\iffalse
\bibliography{ref/refs}
\fi

\section{Conclusions}

In this paper, we proposed a RL-based negotiation-aware adaptive motion planning framework to deal with the sidewalk problem by alleviating social dilemmas that exist in autonomous driving. We employ ACKTR and curriculum learning to train the policy network with sequential OGM images as inputs. According to the prediction horizon policy generated by the policy network, we use polynomial planner with prediction module to generating the optimal trajectories. The effectiveness of our framework is evaluated both in simulation and real-driving scenarios. We find that our framework could regulating driving social behavior and outperform the common alternatives in driving efficiency with safety properly considered.
%%%%%%%%%%%%%%%%%%%%%%%%%%%%%%%%%%%%%%%%%%%%%%%%%%%%%%%%%%%%%%%%%%%%%%%%%%%%%%%%%
\balance
% \addtolength{\textheight}{-12cm}   % This command serves to balance the column lengths

%%%%%%%%%%%%%%%%%%%%%%%%%%%%%%%%%%%%%%%%%%%%%%%%%%%%%%%%%%%%%%%%%%%%%%%%%%%%%

% \section*{Acknowledgement}

% The authors would like to thank

%%%%%%%%%%%%%%%%%%%%%%%%%%%%%%%%%%%%%%%%%%%%%%%%%%%%%%%%%%%%%%%%%%%%%%%%%%%%%

\bibliographystyle{ieeetr}
% \bibliography{section/ref/refs}

\end{document}